\documentclass{tlp}

\usepackage[english]{babel}
\usepackage{latexsym}

\usepackage{epsfig}
\usepackage{url}
\usepackage{rotating}

\newif\ifmakebbl

\newtheorem{definition}{Definition}
\newtheorem{theorem}{Theorem}
\newtheorem{proposition}{Proposition}
\newtheorem{corollary}{Corollary}
\newtheorem{lemma}{Lemma}
\newtheorem{example}{Example}

\newcounter{myenumctr}

\newcommand{\la}{\leftarrow}
\newcommand{\ra}{\rightarrow}
\newcommand{\naf}{{\it not}\,}

\newcommand{\AS}{\mathcal{AS}}

\newcommand{\equivs}{\equiv}

\newcommand{\Pol}{{\rm P}}

\newcommand{\NP}{\mbox{\rm NP}}

\newcommand{\CONP}{\mbox{\rm coNP}}

\newcommand{\SigmaP}[1]{{\Sigma}_{#1}^{P}}
\newcommand{\PiP}[1]{{\Pi}_{#1}^{P}}

\newcommand{\iec}[0]{i.e.,\ }

\newcommand{\commadots}[0]{,\ldots ,}

\renewcommand{\O}{{\cal O}}
\renewcommand{\H}{{\cal H}}
\newcommand{\B}{{\cal B}}

\newcommand{\U}{{\cal U}}
\newcommand{\C}{{\cal C}}
\newcommand{\class}[1]{\C_{\langle #1 \rangle}}
\newcommand{\CU}{\C_\U}

\newcommand{\A}{{\cal A}}
\newcommand{\IU}{2^{{\cal U}}}

\renewcommand{\t}[1]{\langle #1 \rangle}

\title[A Common View on Strong, Uniform, and Other Notions of Equivalence]%
{A Common View on Strong, Uniform, and Other Notions of Equivalence
in Answer-Set Programming\thanks{%
A preliminary version of this paper appeared
in the Proceedings of the 
LPNMR'07 Workshop on Correspondence and Equivalence for Nonmonotonic Theories,
        Vol.\ 265 of the 
CEUR Workshop Proceedings,
2007.}}
\author[S.\ Woltran]
{STEFAN WOLTRAN \\
           Technische Universit\"{a}t
           Wien,
Institut f{\"u}r Informationssysteme
184/2, \\
Favoritenstrasse 9-11,  A-1040 Vienna, Austria\\
           \email{woltran@dbai.tuwien.ac.at}}

\submitted{25 June 2007}
\revised{29 November 2007}
\accepted{5 December 2007}

\begin{document}

\maketitle

\begin{abstract}
Logic programming under the answer-set semantics nowadays deals with
numerous different notions of program equivalence. This is due to the 
fact that equivalence for substitution (known as strong equivalence) 
and ordinary equivalence are different concepts.  The former holds, 
given programs $P$ and $Q$, iff $P$ can be faithfully replaced by $Q$ 
within any context $R$, while the latter holds iff $P$ and $Q$ provide 
the same output, that is, they have the same answer sets.  
Notions in between strong and ordinary equivalence have been introduced 
as theoretical tools to compare incomplete programs and are defined by 
either restricting the syntactic structure of the considered context 
programs $R$ or by bounding the set $\A$ of atoms allowed to occur in $R$ 
(relativized equivalence). For the latter approach, different $\A$ yield 
properly different equivalence notions, in general. For the former 
approach, however, it turned out that any ``reasonable'' syntactic 
restriction to $R$ coincides with either ordinary, strong, or uniform
equivalence (for uniform equivalence, the context ranges over arbitrary 
sets of facts, rather than program rules).  In this paper, we propose 
a parameterization for equivalence notions which takes care of both such 
kinds of restrictions simultaneously by bounding, on the one hand, the 
atoms which are allowed to occur in the rule heads of the context and, on 
the other hand, the atoms which are allowed to occur in the rule bodies 
of the context.  We introduce a general semantical characterization which 
includes known ones as SE-models (for strong equivalence) or UE-models 
(for uniform equivalence) as special cases.  Moreover, we provide 
complexity bounds for the problem in question and sketch a possible 
implementation method making use of dedicated systems for checking ordinary 
equivalence.
\end{abstract}


\section{Introduction}
Starting with the seminal paper on strong equivalence between logic programs
by Lifschitz, Pearce, and Valverde \shortcite{Lifschitz01}, 
a new research direction in 
logic programming under the
answer-set semantics 
has been established.
This is due to fact that strong equivalence
between programs $P$ and $Q$, which holds iff 
$P$ can faithfully be replaced by $Q$ in any program, 
is a different concept than 
deciding whether $P$ and $Q$ have the same answer sets, \iec
whether (ordinary) equivalence between $P$ and $Q$ holds.
Formally, $P$ and $Q$ are strongly equivalent iff, for each further 
so-called context program $R$, 
$P\cup R$ and $Q\cup R$ possess the same answer sets.
That difference between strong and ordinary equivalence 
motivated 
investigations of equivalence notions in between (see \cite{Eiter07} for an overview).
Such equivalence notions were obtained in 
two basic ways, viz.\ to 
bound the actually allowed context programs $R$ by 
(i) restricting their syntax; or 
(ii) restricting their alphabet. 

For Case~(i), it turned out that any ``reasonable''
(\iec where the restriction is defined rule-wise, for instance only
allowing context programs with Horn rules)
attempt
coincides with either ordinary, strong,
or uniform equivalence (see, e.g., \cite{Pearce04}).
The later notion,
uniform equivalence, was originally introduced by Sagiv~\shortcite{sagi-88} as
an approximation for
datalog equivalence and has been 
adapted to answer-set programming by Eiter and Fink~\shortcite{eite-fink-03}.
Uniform equivalence
tests whether,
for 
each set $F$ of facts, $P\cup F$ and $Q\cup F$ possess the same answer sets.
Case~(ii),
where the atoms allowed to occur in $R$ are 
from a given 
alphabet $\A$ yields in general different concepts for 
different $\A$ and thus is known as
strong equivalence relative to $\A$~\cite{Woltran04}.
A combination of both approaches
leads to the concept of uniform equivalence relative to $\A$~\cite{Woltran04}.%
\footnote{A further direction of research is to additionally restrict the
alphabet over which the answer sets of 
$P\cup R$ and $Q\cup R$ are
compared. This kind of \emph{projection} was investigated 
in~\cite{Eiter05,OJ06:ecai,Oetsch07b}, but is beyond the scope of this work.}

In this paper, we propose a 
framework to 
define
more
fine-grained 
notions of equivalence, such 
that  
the
aforementioned restrictions are 
captured
simultaneously. 
This is accomplished
by parameterizing,
on the one hand,
the atoms which are allowed to occur in the 
rule heads of the context programs
and, on the other hand,
the atoms which are allowed to occur in the 
rule bodies of  the context programs.
More formally, the problem we study is as follows, and we will refer to it 
as $\t{\H,\B}$-equivalence:
\begin{quote}
Given programs $P$, $Q$, 
and alphabets $\H$, $\B$, 
decide whether the answer sets of $P\cup R$ and $Q\cup R$ coincide
for each program $R$, where each rule in $R$ has its head atoms from $\H$ 
and its body atoms from $\B$.
\end{quote}
As we will show, for all such kinds of equivalence
it is safe to consider only unary rules 
(that are simple rules of the form $a\la $ or $a\la b$)
in context programs $R$.
Therefore,
instances of $\t{\H,\B}$-equivalence include 
all previously mentioned equivalence notions.
In particular, 
for $\B=\emptyset$, \iec
disallowing any atom to occur in bodies,
our notion amounts to
uniform equivalence relative to $\H$.
Moreover,
the parameterization
$\H=\B$ amounts to relativized strong equivalence. 
As a consequence, we obtain 
(unrelativized) strong equivalence
if
$\H=\B=\U$, where $\U$ is the universe  of atoms, 
and 
(unrelativized) uniform equivalence
if $\H=\U$ and $\B=\emptyset$.

The main contribution of the paper is
to provide a general  and uniform
semantic characterization for 
the newly introduced 
framework.
Moreover, we show 
that our 
characterization includes as special cases 
prominent ones 
for strong 
and uniform equivalence, namely the so-called
SE-models due to Turner~\shortcite{Turner03}, and respectively, the
so-called 
UE-models due to Eiter and Fink~\shortcite{eite-fink-03}, and 
thus clarifies the
differences which have been observed between these known characterizations.
As well, the relativized variants of SE-models and UE-models 
\cite{Woltran04} will be shown to be special cases of our new
characterization.
Finally, we 
address the computational complexity of the
decision problems for the 
introduced equivalence notions.
The complexity results suggest to implement 
tests for $\t{\H,\B}$-equivalence via existing dedicated systems for checking ordinary equivalence.
We briefly sketch such a method at the end of the paper.

The benefits of the introduced framework are twofold. 
On the one hand, we provide a unified method to decide different notions
of equivalence by the same concept. So far, such methods were conceptually different
for strong and uniform equivalence, and thus our results shed new light on the 
origin of these differences.
On the other hand, the introduced equivalence notion allows 
to precisely specify in which scenarios a program $P$ can 
be replaced by a potential simplification $Q$. 
For instance, suppose a program $P$ is given over atoms $\U$ and provides an output 
over atoms $\O\subseteq\U$. These output atoms
are only used in rule bodies of 
potential extensions of $P$, whereas all other atoms can be used arbitrarily in such
extension.
In such a scenario, $Q$ can faithfully be used as a simplification of $P$, 
in case $\t{\H,\B}$-equivalence 
between $P$ and $Q$ with
parameters $\H=\U\setminus \O$
and $\B=\U$ holds.

\section{Background}
Throughout the paper we assume an arbitrary finite but fixed
universe $\U$ of atoms. 
Subsets of the universe are either called interpretations or 
alphabets: We use the latter term to restrict the syntax of programs, while
the former is used in the context of semantics for programs.
For an interpretation $Y$ and an alphabet $\A$, we write
$Y|_\A$ instead of $Y\cap \A$.

A propositional disjunctive logic program (or simply, a program)
is a finite set of rules of the form 
\begin{equation}\label{eq:r} 
a_1\vee \cdots \vee a_l \la a_{l+1},\ldots,a_m,\naf
 a_{m+1},\ldots,\naf a_n,
\end{equation}
($n>0$,
$n\geq m \geq l$), 
and
where all $a_i$ 
are propositional atoms from $\U$ and $\naf$  denotes
default negation; for $n=l=1$, 
we usually identify the rule (\ref{eq:r}) 
with the atom $a_1$, and call it 
a \emph{fact}.
A rule of the form 
(\ref{eq:r}) 
is called 
a \emph{constraint} if $l=0$;
\emph{positive} if $m=n$;
\emph{normal} if $l\leq 1$; 
and \emph{unary} if it 
is
either a fact or of the form $a\la b$. 
A program is positive (resp., normal, 
unary) iff all its rules
are positive (resp., normal, 
unary).
If all atoms occurring in a program $P$ are from a given alphabet 
$\A\subseteq \U$ 
of atoms, we say that $P$ is a program \emph{over} (alphabet) $\A$.
The class of all logic programs 
(over the fixed universe $\U$) is denoted by $\CU$.

For a rule $r$ of the form (\ref{eq:r}), we identify its 
head by $H(r) = \{ a_1\commadots a_l \}$ and its
body via 
$B^+(r) = \{a_{l+1},\ldots,a_m \}$ and
$B^-(r) = \{a_{m+1},\ldots,a_n \}$. 
We shall write rules of the form (\ref{eq:r}) also as
$H(r)\la B^+(r),\naf B^-(r)$.
Moreover,
we use $B(r) = 
B^+(r)\cup B^-(r)$.
Finally, 
for a program $P$
and
$\alpha\in\{H,B,B^+,B^-\}$,
let
$\alpha(P)=\bigcup_{r\in P} \alpha(r)$.

The relation $Y\models r$ between an interpretation $Y$ and a program $r$ is defined  as usual, 
i.e., $Y\models r$ iff 
$H(r)\cap Y\neq\emptyset$, whenever
jointly $B^+(r)\subseteq Y$ and $B^-(r)\cap Y=\emptyset$ hold;
for a program $P$,
$Y\models P$ holds iff for each $r\in P$, 
$Y\models r$. 
If $Y\models P$ holds, $Y$ is 
called a \emph{model} of $P$.
Following Gelfond and Lifschitz~\shortcite{gelf-lifs-91}\footnote{%
However, we omit strong (``classical'') negation here;
our results can be generalized to extended logic programs
the same way as discussed in \cite{Eiter07}.},
an interpretation~$Y$
is an \emph{answer set} of a program $P$ 
iff it is a minimal (w.r.t.\ set inclusion) model of the \emph{reduct} 
$P^Y = \{ H(r) \la B^+(r) \mid Y\cap B^-(r) = \emptyset\}$ 
of $P$ w.r.t.\ $Y$.
The set of all answer sets of a program $P$ is denoted by $\AS(P)$.

Next, we review some prominent notions of equivalence, 
which have
been studied under the answer-set semantics:
Programs $P,Q\in\CU$ are 
\emph{strongly equivalent}~\cite{Lifschitz01}, iff, for any program $R\in\CU$, 
$\AS(P\cup R)=\AS(Q\cup R)$; $P$ and $Q$ are 
\emph{uniformly equivalent}~\cite{eite-fink-03}, iff, for any  set $F\subseteq \U$ of facts,
$\AS(P\cup F)=\AS(Q\cup F)$.
Relativizations of these notions are as follows~\cite{Woltran04,Eiter07}:
For a given alphabet $\A\subseteq \U$, 
we call programs $P,Q\in\CU$
\emph{strongly equivalent relative to} $\A$,
iff, for any program $R$ over $\A$,
it holds that $\AS(P\cup R)=\AS(Q\cup R)$;
$P,Q$ are 
\emph{uniformly equivalent relative to} $\A$,
iff, for any set 
$F\subseteq \A$ of facts,
$\AS(P\cup F)=\AS(Q\cup F)$.
Finally,
if $\A=\emptyset$, we obtain \emph{ordinary equivalence},
\iec $\AS(P)=\AS(Q)$ for both strong and uniform equivalence 
relative to $\A$.

In case of strong equivalence (also in the relativized case),
it was shown~\cite{Lifschitz01,Woltran04}
that the syntactic class of \emph{counterexamples}, \iec
programs $R$, such that $\AS(P\cup R)\neq\AS(Q\cup R)$,
can always be restricted to 
the class of unary programs. 
Hence, the next result comes 
without surprise, but additionally provides 
insight
with respect to the alphabets in the rules' 
heads and bodies. 

\begin{lemma}\label{lemma:1}
For any programs $P$, $R \in  \CU$ and
any interpretation $Y$, 
there exists a positive program $R'$ 
such that 
$H(R')\subseteq H(R)$, $B(R')\subseteq B(R)$, 
and
$Y\in \AS(P\cup R)$ iff $Y\in\AS(P\cup R')$.
\end{lemma}
\begin{proof}
Recall that 
$Y\models P$ iff
$Y\models P^Y$ holds. Moreover, $(P\cup R)^Y = (P\cup R^Y)^Y = (P^Y\cup R^Y)$
is clear.
Thus, $Y\in \AS(P\cup R)$ iff $Y\in \AS(P\cup R^Y)$ is obvious.
By definition
$R^Y$ is positive and 
satisfies $H(R^Y)\subseteq H(R)$, $B(R^Y)\subseteq B(R)$.
Thus the claim follows using $R'=R^Y$.
\end{proof}

As we will see later, Lemma~\ref{lemma:1} can even be strengthened
to a unary program $R'$.
In terms of equivalence checking, Lemma~\ref{lemma:1} 
has some interesting consequences.
First, observe that if two programs 
$P$, $Q$ do not have the same answer
sets, 
the \emph{common} rules $R$ 
from  $P$ and $Q$
can be significantly simplified,
without changing the witnessing answer set $Y$.
Second, in terms of strong equivalence, 
the result shows that whenever a counterexample $R$
for a strong equivalence problem between $P$ and $Q$ exists, 
\iec  $\AS(P\cup R)= \AS(Q\cup R)$ does not hold,
then 
we can find a simpler, in particular, a positive counterexample $R'$, 
which is given over the same alphabets (as $R$) for
heads, and respectively, bodies.
In other words, for deriving proper different equivalence notions, it turns
out that the alphabets
of the atoms allowed to occur in rule heads, 
and respectively,
(positive) rule bodies of the context programs are the 
crucial parameters.

\section{The General Framework}

We now formally ground these considerations and start by
introducing the following classes of logic programs.

\begin{definition}\label{def:class}
For any alphabets $\H,\B\subseteq \U$,
the class $\class{\H,\B}$ of programs is defined
as $\{ P \in \CU \mid H(P)\subseteq \H, B(P)\subseteq \B\}$.
\end{definition}

With this concept of program classes at hand, we define
equivalence notions which are more fine-grained than the 
ones previously discussed.

\begin{definition}
Let $\H,\B\subseteq \U$ be alphabets, and $P,Q\in\CU$ be programs.
The \emph{$\t{\H,\B}$-equivalence problem} between 
$P$ and $Q$, in symbols 
$P\equiv_{\t{\H,\B}} Q$, 
is to decide whether, for each $R\in\class{\H,\B}$,
$\AS(P\cup R)=\AS(Q\cup R)$. 
If $P\equiv_{\t{\H,\B}} Q$ holds, we 
say that $P$ and $Q$ are \emph{$\t{\H,\B}$-equivalent}.
\end{definition}

The class
$\class{\H,\B}$ is also called the 
\emph{context} of an $\t{\H,\B}$-equivalence problem, 
and a program $R\in \class{\H,\B}$, such that
$\AS(P\cup R) = \AS(Q\cup R)$ does not hold, is called a 
\emph{counterexample} to the $\t{\H,\B}$-equivalence problem
between
$P$ and $Q$.

\begin{example}\label{ex:1}
Consider programs
$$
P= \{ a \vee b \la;\;a \la b \}
\quad\mbox{and}\quad
Q = \{ a \la \naf b;\; b\la\naf a;\; a \la b \}.
$$
It is known that these programs are not strongly equivalent, since
adding any $R$ which closes the cycle between $a$ and $b$
yields $\AS(P\cup R)\neq\AS(Q\cup R)$.
In particular, for $R=\{b\la a\}$, we get $\AS(P\cup R)=\{\{a,b\}\}$, while
$\AS(Q\cup R)=\emptyset$.
However, $P$ and $Q$ are uniformly equivalent.
In terms of $\t{\H,\B}$-equivalence
we are able to ``approximate''
equivalence notions which hold between $P$ and $Q$.
It can be shown that, 
for instance,
$P \equiv_{\t{\{a,b\},\{b\}}} Q$ 
or
$P \equiv_{\t{\{a\},\{a,b\}}} Q$ 
holds
(basically since $b\la a$ does not occur in any program in
$\C_{\t{\{a,b\},\{b\}}}$, 
or
$\C_{\t{\{a\},\{a,b\}}}$).
On the other hand, 
$P \not\equiv_{\t{\{b\},\{a,b\}}} Q$,
and likewise,
$P \not\equiv_{\t{\{a,b\},\{a\}}} Q$,
since $\{b\la a\}$ is 
contained in the context 
$\C_{\t{\{b\},\{a,b\}}}$,
resp., 
in
$\C_{\t{\{a,b\},\{a\}}}$.
\hfill$\diamond$
\end{example}

Observe that the concept of $\t{\H,\B}$-equivalence
captures other equivalence notions as follows:
$\langle \A,\A \rangle$-equivalence coincides with strong equivalence relative to $\A$; and,
in particular, $\langle \U,\U \rangle$-equivalence 
coincides with strong 
equivalence.
We will show that 
$\langle \A,\emptyset \rangle$-equivalence amounts to 
uniform equivalence relative to $\A$;
and, in particular,
$\langle \U,\emptyset \rangle$-equivalence
amounts to uniform  
equivalence.%
\footnote{
For a graphical illustration of 
different parameterizations of
$\t{\H,\B}$-equivalence
with respect to those special cases, 
see Figure~\ref{fig:1} in the conclusion.}
Note that the relation to uniform equivalence is not immediate
since $\langle \A,\emptyset \rangle$-equivalence deals
with a context containing 
sets of \emph{disjunctive} facts, \iec rules of the form
$a_1\vee\cdots\vee a_l\la$, rather than sets of (simple) facts, \iec
rules of the form $a\la$.

A central aspect 
in equivalence checking is the 
quest for semantical characterizations which are 
assigned
to a \emph{single} program. 
In particular, this is vital if a program is 
compared to numerous other programs, which, for instance,
are considered as possible
candidates for optimizations.

\begin{definition}\label{def:sc}
A \emph{semantical characterization} for an 
$\t{\H,\B}$-equivalence problem is understood as
a function 
$\sigma_{\t{\H,\B}}: \CU \ra 2^{\IU \times \IU}$
mapping each program to a set of pairs of interpretation,
such that,
for any 
$P,Q\in\CU$, 
$P\equiv_{\t{\H,\B}} Q$ holds iff
$\sigma_{\t{\H,\B}}(P)=\sigma_{\t{\H,\B}}(Q)$.
\end{definition}
 
Since we are interested in a uniform characterization 
of $\t{\H,\B}$-equivalence problems, we in fact provide 
a single function $\sigma:\U\times\U\times\CU\ra 2^{\IU \times \IU}$,
such that for any programs $P,Q\in\CU$ and any 
alphabets $\H,\B\subseteq\U$, 
$P\equiv_{\t{\H,\B}} Q$ holds iff
$\sigma(\H,\B,P)=\sigma(\H,\B,Q)$.
However, for the sake of uniformity we will
use $\sigma_{\t{\H,\B}}(P)$ and
$\sigma(\H,\B,P)$ interchangeably.

We will review known semantical characterizations for 
special cases 
(as, for instance, SE-models~\cite{Turner03} 
and UE-models~\cite{eite-fink-03} for 
strong, and respectively, uniform equivalence) later. 

Finally, we introduce 
containment problems.

\begin{definition}
Let $\H, \B\subseteq \U$ be alphabets, and $P,Q\in\CU$ be programs.
The \emph{$\t{\H,\B}$-containment problem} for 
$P$ in $Q$,
in symbols 
$P\subseteq_{\t{\H,\B}} Q$,
 is to decide whether, for each 
$R\in\class{\H,\B}$,
$\AS(P\cup R)\subseteq \AS(Q\cup R)$. 
A counterexample to $P\subseteq_{\t{\H,\B}} Q$, is any
program $R\in \class{\H,\B}$, such that 
$\AS(P\cup R)\not\subseteq \AS(Q\cup R)$.
\end{definition}

Containment and equivalence problem are closely related by definition:

\begin{proposition}\label{prop:eqcont}
For any programs $P,Q\in\CU$ and any alphabets $\H,\B\subseteq\U$,
$P\equiv_{\t{\H,\B}} Q$ holds iff
$P\subseteq_{\t{\H,\B}} Q$ and 
$Q\subseteq_{\t{\H,\B}} P$ jointly hold.
\end{proposition}

\section{Characterizations for $\t{\H,\B}$-Equivalence}

Towards the semantical characterization for 
$\t{\H,\B}$-equivalence problems, we first
introduce the notion of a witness, 
which is assigned to $\t{\H,\B}$-containment problems
taking both compared programs into 
account. 
Afterwards, we will derive the desired semantical characterization
of $\t{\H,\B}$-models which are
assigned to single programs 
along the lines of Definition~\ref{def:sc}.

Before that, we need some further technical concepts and results.

\begin{definition}
Given alphabets $\H,\B\subseteq \U$, we define the relation
$\preceq_\H^\B\subseteq \U\times\U$ between interpretations as follows:
$V\preceq_\H^\B Z$ iff
$V|_\H\subseteq Z|_\H$ and $Z|_\B\subseteq V|_\B$.
\end{definition}

Observe that
if $V\preceq_\H^\B Z$ holds, then
either 
$V|_{\H\cup B}=Z|_{\H\cup B}$ holds, 
or at least one out  of
$V|_\H\subset Z|_\H$ and $Z|_\B\subset V|_\B$ holds.
We write $V\prec_\H^\B Z$, in case
$V\preceq_\H^\B Z$ and $V|_{\H\cup B}\neq Z|_{\H\cup B}$.
Observe that 
$V\prec_\H^\B Z$ thus holds 
iff 
$V\preceq_\H^\B Z$ and 
$Z\not\preceq_\H^\B V$
jointly hold.

\begin{lemma}\label{lemma:2}
Let $\H,\B\subseteq \U$ be alphabets,
$P$ a positive program 
with $H(P)\subseteq \H$, $B(P)\subseteq \B$,
and $Z,V\subseteq \U$ interpretations.
Then, 
$V\models P$ and $V\preceq_\H^\B Z$
imply
$Z\models P$. 
\end{lemma}
\begin{proof}
Suppose
$Z\not\models P$  and
$V\preceq_\H^\B Z$, \iec
$V|_\H \subseteq Z|_\H$ and
$Z|_{\B}  \subseteq V|_{\B}$ hold.
We show $V\not\models P$.
If $Z\not\models P$, then 
there exists a rule $r\in P$, 
such that $B^+(r)\subseteq Z$ and $Z\cap H(r)=\emptyset$.
Since $H(r)\subseteq \H$, we get from $V|_\H\subseteq Z|_\H$, that 
$V\cap H(r)=\emptyset$. 
Moreover, since 
$B^+(r)\subseteq \B$, 
we have $B^+(r)\subseteq Z|_{\B} \subseteq V|_{\B}$, and thus
$B^+(r)\subseteq V$. 
Hence $V\not\models r$ which yields $V\not\models P$. 
\end{proof}

We also need the concept of an $\H$-total model.

\begin{definition}
Given $\H\subseteq \U$, an interpretation $Y$
is an \emph{$\H$-total model} for a program $P\in\CU$ iff
$Y\models P$ and, for all $Z\subset Y$, 
$Z\models P^Y$ implies $Z|_\H\subset  Y|_\H$.
\end{definition}

$\H$-total models of a program $P$ are the only ones which can be 
turned into an answer set by adding a program $R\in\class{\H,\B}$ to $P$.

\begin{lemma}\label{lemma:htotal}
Let $P\in\CU$ be a program, $Y$ be an interpretation
and $\H,\B\subseteq \U$ be alphabets.
Then, there exists a program $R\in \class{\H,\B}$, such 
that $Y\in\AS(P\cup R)$ only if
$Y$ is an $\H$-total model of $P$.
\end{lemma}
\begin{proof}
If there exists a program $R\in \class{\H,\B}$, such
that $Y\in\AS(P\cup R)$ then, by Lemma~\ref{lemma:1} there is also
a positive program $R'\in\class{\H,\B}$, such that 
$Y\in\AS(P\cup R')$. Hence, $Y\models P\cup R'$ holds. 
From $Y\models R'$, we get for all $Z$ with $Y\preceq^\B_\H Z$, 
$Z\models R$. 
This includes in particular all $Z\subset Y$ with 
$Z|_\H=Y|_\H$. Hence, for each 
such $Z$, $Z\not\models P^Y$ has to hold, otherwise
$Y\in\AS(P\cup R')$ would not hold.
But then, $Y$ is an $\H$-total model of $P$ by definition.
\end{proof}

\subsection{Witnesses for Containment Problems}

In order to find a counterexample for an $\t{\H,\B}$-inclusion problem
$P\subseteq_{\t{\H,\B}} Q$, 
we thus need on the one hand an $\H$-total model of $P$ but in addition
we need to take further conditions for $Q$ into account. 
This is captured by the following concept.

\begin{definition}\label{def:witness}
A \emph{witness} against a containment problem 
$P\subseteq_{\t{\H,\B}} Q$ is a pair of interpretations
$(X,Y)$ with $X\subseteq Y\subseteq \U$, such that
\begin{itemize}
\item [(i)] 
$Y$ is an $\H$-total model of $P$; and
\item [(ii)] 
if $Y\models Q$ then $X\subset Y$, 
$X\models Q^Y$, and 
for each $X'$
with $X\preceq_\H^\B X' \subset Y$,
 $X'\not \models P^Y$. 
\end{itemize}
\end{definition}

We 
prove that the existence of witnesses against
a containment problem $P\subseteq_{\t{\H,\B}} Q$
shows that  $P\subseteq_{\t{\H,\B}} Q$ does not hold. 
As a by-product we obtain that 
there are always counterexamples to 
$P\subseteq_{\t{\H,\B}} Q$ 
of a simple syntactic form.

\begin{lemma}\label{lemma:witness}
The following propositions are equivalent for any 
$P,Q\in\CU$ and any $\H,\B\subseteq \U$:
\begin{itemize}
\item[(1)]
$P\subseteq_{\t{\H,\B}} Q$ 
does not hold;
\item[(2)]
there exists a unary program $R\in\class{\H,\B}$, such 
that $\AS(P\cup R)\not\subseteq\AS(Q\cup R)$;
\item[(3)]
there exists a witness against
 $P\subseteq_{\t{\H,\B}} Q$.
\end{itemize}
\end{lemma}
\begin{proof}
We show that (1) implies (3) and (3) implies (2). 
(2) implies (1) 
obviously holds by definition of $\t{\H,\B}$-containment problems. 

(1) implies (3): If $P\subseteq_{\t{\H,\B}} Q$ does not hold,
there exists a program $R$, and an interpretation $Y$, such 
that $Y\in\AS(P\cup R)$ and $Y\notin\AS(Q\cup R)$. 
By Lemma~\ref{lemma:htotal}, 
$Y$ has to be an $\H$-total of $P$.
It remains to establish Property~(ii) in Definition~\ref{def:witness}.
From $Y\notin\AS(Q\cup R)$, we either get $Y\not\models Q\cup R$ or 
existence of an $X\subset Y$ such that $X\models (Q\cup R)^Y$.
Recall that
by Lemma~\ref{lemma:1},
we can w.l.o.g.\ assume that $R$ is positive;
thus, $(Q\cup R)^Y=(Q^Y\cup R)$
We already know that $Y\models R$
(otherwise $Y\in\AS(P\cup R)$ cannot hold). 
Hence,
in the former case, \iec $Y\not\models Q\cup R$, we get
$Y\not\models Q$.
Then, for any $X\subseteq Y$, $(X,Y)$ is a witness 
against $P\subseteq_{\t{\H,\B}} Q$, and we are done.
For the remaining case, 
where $X\models Q^Y$ and $X\models R$,
we suppose
towards a contradiction, 
that there exists an $X'\subset Y$, such that 
$X'\models P^Y$ and
$X\preceq_\H^\B X'$ hold.
The latter together with $X\models R$ yields
$X'\models R$, following Lemma~\ref{lemma:2}.
Together with $X'\models P^Y$, we thus get 
$X'\models (P^Y\cup R)=(P\cup R)^Y$. 
Since $X'\subset Y$ this is in contradiction to $Y\in\AS(P\cup R)$.
Thus $(X,Y)$ is a witness against 
$P\subseteq_{\t{\H,\B}} Q$. 
\smallskip

\noindent
(3) implies (2): 
Let $(X,Y)$ be a witness 
against $P\subseteq_{\t{\H,\B}} Q$.
We use the unary program
$$
R = X|_\H \cup \{ a \la b \mid a \in (Y\setminus X)|_\H, b\in (Y\setminus X)|_\B \}
$$
and show 
$Y\in\AS(P\cup R)\setminus AS(Q\cup R)$. 
Note that $R$ is 
contained in class 
$R\in\class{\H,\B}$, since
the set $X|_\H\subseteq\H $ of facts uses only atoms from $\H$, and
all further rules $a\la b$ in $R$  satisfy $a\in \H$ and $b\in \B$ by definition. 
We first show $Y\in\AS(P\cup R)$. Since $(X,Y)$ is a witness 
against $P\subseteq_{\t{\H,\B}} Q$,
we know
$Y\models P$. $Y\models R$ is easily checked and thus
$Y\models P\cup R$.
It remains to show that no 
$Z\subset Y$ satisfies $Z\models (P\cup R)^Y=P^Y \cup R$.
Towards a contradiction suppose such a $Z$ exists.
Hence, $Z\models P^Y$ and $Z\models R$.
From $Z\models R$, we get that $X|_\H\subseteq Z|_\H$ has to hold.
Since $(X,Y)$ is a witness against 
$P\subseteq_{\t{\H,\B}} Q$, 
$Z|_\H \subset Y|_\H$ holds, 
since $Y$ has to an $\H$-total model of $P$, following
Definition~\ref{def:witness}.
Hence, $X|_\H\subseteq Z|_\H \subset Y|_\H$ holds. 
We have $X\subset Y$
and, moreover, get 
$Z|_\B \not \subseteq X|_\B$
from Property~(ii) in Definition~\ref{def:witness},
since 
$Z\models P^Y$ and $X|_\H\subseteq Z|_\H$ already hold.
Now,
$Z|_\B\subseteq Y|_\B$ follows from our assumption $Z\subset Y$, 
hence there exists an atom $b\in (Y\setminus X)|_\B$ contained in $Z$.
We already know that $X|_\H \subseteq Z|_\H \subset Y|_\H$ has to hold.
Hence, there exists at least one $a\in (Y\setminus X)|_\H$, not
contained in $Z$.
But then, we derive that $Z\not\models 
\{ a \la b \}$. 
Since $a \la b \in R$, this is
a contradiction to $Z\models R$.

It remains to show $Y\notin\AS(Q\cup R)$.
If $Y\not\models Q$, we are done. So let $Y\models Q$. 
Since $(X,Y)$ is a witness against $P\subseteq_{\t{\H,\B}} Q$, 
we get $X\models Q^Y$ and $X\subset Y$.
It is easy to see that $X\models R$ holds.
Thus $X\models (Q^Y\cup R)=(Q\cup R)^Y$, and 
$Y\notin\AS(Q\cup R)$ follows.
\end{proof}

We illustrate how to obtain witnesses on some examples.

\begin{example}\label{ex:2}
We already have mentioned 
in Example~\ref{ex:1}
that 
$$
P= \{ a \vee b \la;\;a \la b \}
\quad\mbox{and}\quad
Q = \{ a \la \naf b;\; b\la\naf a;\; a \la b \}
$$
are not $\t{\H,\B}$-equivalent
for $\H=\{b\}$ and 
$\B=\{a,b\}$.
We show that there exists a witness against $P\subseteq_{\t{\H,\B}} Q$.

We start with the models (over $\{a,b\}$) of $P$, which are
$$
Y_1 = \{a,b\}\quad\mbox{and}\quad Y_2=\{ a\}.
$$
Both are also $\H$-total models of $P$, and moreover, 
$\H$-total
models of $Q$.
For $Y_1$ this is the case since $\{b\}\not\models P^{Y_1}=P$ and 
$\{b\}\not\models Q^{Y_1}=\{a\la b\}$. 
For $Y_2$ we have 
$\emptyset\not\models P^{Y_2}=P$ and
$\emptyset\not\models Q^{Y_2}=\{a;\;a\la b\}$.
Now, in order to find a witness against
$P\subseteq_{\t{\H,\B}} Q$
we need to find for some $i\in\{1,2\}$ an interpretation $X_i\subset Y_i$, such 
that $X_i\models Q^{Y_i}$ and for each $X'$ with $X_i\preceq^\B_\H X'\subset Y_i$, 
$X'\not\models P^{Y_i}$.

Let us use $i=1$. The models 
of $Q^{Y_1}$ 
which are a proper subset of $Y_1$ are 
$\emptyset$ and $\{a\}$.
Let $X_1=\emptyset$. It remains to check that for each $X'$ with
$X_1\preceq^\B_\H X' \subset Y_1$, $X'\not\models P^{Y_1}$.
Since $\B=\{a,b\}$ 
$X_1\preceq^\B_\H X'$ implies
that $X'|_\B\subseteq  X_1|_\B$, 
\iec $X'\subseteq X_1$ has to hold.
Hence, 
the only $X'$  (over $\{a,b\}$) 
satisfying $X_1\preceq^\B_\H X'$
is $X_1$ itself.
Since $P^{Y_1}=P$, we have $X_1\not\models P^{Y_1}$ and 
thus $(X_1,Y_1)$ is a witness against $P\subseteq_{\t{\H,\B}} Q$.
One can check that 
this is in fact the only witness (over $\{a,b\}$) against
$P\subseteq_{\t{\H,\B}} Q$.

We also have mentioned
in Example~\ref{ex:1}
that 
$P$ and
$Q$ 
are not $\t{\H',\B'}$-equivalent
for
$\H'=\{a,b\}$ and $\B'=\{a\}$.
Let us again find a witness against the containment problem
$P\subseteq_{\t{\H',\B'}} Q$.
Again $Y_1$ and $Y_2$ 
as above
are $\H'$-total models (which is here easy to see, since 
$\H'$ is now the universe $\{a,b\}$). 
Let us check whether $(X_1,Y_1)$ with $X_1=\emptyset$ is 
now also a witness against $P\subseteq_{\t{\H',\B'}} Q$.
The argumentation is slightly different, 
for $\B'=\{a\}$:
in fact,
we now have two candidates for $X'$ to satisfy
$X_1\preceq^{\B'}_{\H'} X'\subset Y_1$, 
viz.\ $X'_1= \emptyset$ and $X'_2=\{b\}$.
However neither of them is a model of $P^{Y_1}=P$ thus
$(X_1,Y_1)$ is also a witness against $P\subseteq_{\t{\H',\B'}} Q$.
\hfill$\diamond$
\end{example}

As an immediate consequence of Lemma~\ref{lemma:witness} and Proposition~\ref{prop:eqcont}, we get that $\t{\H,\B}$-equivalence
problems which do not hold always possess simple counterexamples.
As a special case we obtain 
the already mentioned fact
that $\t{\H,\emptyset}$-equivalence amounts to
uniform equivalence relative to $\H$, 
and, in particular, $\t{\U,\emptyset}$-equivalence coincides 
with the notion of uniform equivalence.
In other words, proper 
disjunctive facts are not of relevance for deciding 
$\t{\H,\emptyset}$-equivalence problems.

\begin{corollary}\label{cor:unary}
For any alphabets $\H,\B\in\U$ and any programs $P,Q\in\CU$, 
$P\equiv_{\t{\H,\B}} Q$ does not hold iff there exists a unary 
program $R\in\class{\H,\B}$, such that
$\AS(P\cup R)\neq\AS(Q\cup R)$; 
if $\B=\emptyset$, then $P\equiv_{\t{\H,\B}} Q$ does not hold iff there exists a
set $F\subseteq \H$ of facts, 
such that $\AS(P\cup F)\neq\AS(Q\cup F)$.
\end{corollary}

\subsection{Introducing $\t{\H,\B}$-models}

Next, we present the desired 
semantical characterization for $\t{\H,\B}$-equivalence, which 
we 
call
$\t{\H,\B}$-models. We need a further formal concept first.

\begin{definition}\label{def:max}
Given $\H,\B\subseteq \U$, a  
pair $(X,Y)$ of interpretations
is called $\preceq^\B_\H$-maximal for $P$ iff
$X\models P^Y$ and, 
for each 
$X'$ with
$X\prec^\B_\H X'\subset Y$,
$X'\not\models P^Y$.
\end{definition}

Observe
that
being
$\preceq^\B_\H$-maximal
refers to being
maximal (w.r.t.\ subset inclusion) in the atoms from $\H$ 
and simultaneously 
minimal (w.r.t.\ subset inclusion) 
in the atoms from $\B$.

\begin{definition}\label{def:model}
Given $\H,\B\subseteq \U$, and interpretations 
$X\subseteq Y\subseteq \U$,
a pair $(X,Y)$ is 
an \emph{$\t{\H,\B}$-model} of a program $P\in\CU$ iff
$Y$ is an $\H$-total model for $P$ and, 
if $X\subset Y$,
there exists an $X'\subset Y$ with $X'|_{\H\cup \B}=X$, such
that
$(X',Y)$ is $\preceq^\B_\H$-maximal for $P$.
 
The set of all 
$\t{\H,\B}$-models of a program $P$ is 
denoted by 
$\sigma_{\langle\H,\B\rangle}(P)$.
\end{definition}

Moreover, let us call a pair
$(X,Y)$ \emph{total} 
if $X=Y$, otherwise it is called \emph{non-total}.
Observe that each non-total 
$\t{\H,\B}$-model $(X,Y)$ satisfies 
$X\subseteq Y|_{\H\cup \B}$ and
$X|_\H\subset Y|_\H$.
The latter comes from the fact that $Y$ is $\H$-total, while
the former holds in view of the conditions $X'|_{\H\cup \B} =X$  
and $X\subset Y$ in the definition.
The reason for that different 
realization of the two interpretations in a non-total $\t{\H,\B}$-model $(X,Y)$ 
of a program $P$
can be briefly explained as follows: 
First, the standard interpretation $Y$ refers to a potential answer set candidate, i.e., 
an interpretation which can be turned into an answer set by adding
a program from $\class{\H,\B}$ to $P$ (see also Lemma~\ref{lemma:htotal}).
Second, the restriction for $X$ to be a subset of $\H\cup \B$ 
is due to the fact of the restricted ``power''
of the programs in $\class{\H,\B}$. 
In fact, suppose we have different models $X'$, $X''$ of the reduct $P^Y$ 
such that $X'|_{\H\cup\B}=X''|_{\H\cup \B}=X$. Then, no matter which $R\in\class{\H,\B}$ is chosen to 
be added to $P$, we either have that both $X'$ and $X''$ are models of $(P\cup R)^Y$ or
neither of them.
Therefore two such models are
collected into the single $\t{\H,\B}$-model $(X,Y)$.

Before stating our main theorem,
we require one further lemma.

\begin{theorem}\label{thm:central}
For any programs $P,Q\in\CU$ and any alphabets $\H,\B\subseteq\U$,
$P \equiv_{\t{\H,\B}} Q$ holds iff
$\sigma_{\t{\H,\B}}(P) = \sigma_{\t{\H,\B}}(Q)$.
\end{theorem}

\begin{proof}
If-direction: Suppose that either $P \subseteq_{\t{\H,\B}} Q$ 
or 
$Q \subseteq_{\t{\H,\B}} P$ 
does not hold.
Let us w.l.o.g.\
assume 
$P \subseteq_{\t{\H,\B}} Q$ does not hold (the other case is symmetric).
By Lemma~\ref{lemma:witness}, then a witness $(X,Y)$
against $P \subseteq_{\t{\H,\B}} Q$ exists. 
By definition of a witness, $Y$ is then an $\H$-total model of $P$ and
we have $(Y,Y)\in\sigma_{\t{\H,\B}}(P)$. In case
$(Y,Y)\notin\sigma_{\t{\H,\B}}(Q)$ we are already done.
So suppose $(Y,Y)\in\sigma_{\t{\H,\B}}(Q)$. Then,
$Y$ has to be $\H$-total for $Q$ as well, 
and we obtain that $X|_\H\subset Y|_\H$,
$X\models Q^Y$, and for each $X'$ 
with $X\preceq^\B_\H X' \subset Y$, $X'\not\models P^Y$ hold.
Consider now a pair
$(Z,Y)$ of interpretations
with $Z\subset Y$
which is 
$\preceq^\B_\H$-maximal for $Q$.
Then $X\preceq^\B_\H Z$ has 
to hold
and
we obtain
$(Z|_{\H\cup \B},Y)\in\sigma_{\t{\H,\B}}(Q)$.
On that other hand, we have
$(Z|_{\H\cup \B},Y)\notin\sigma_{\t{\H,\B}}(P)$.
This is a consequence of the observation that 
for each $X'$
with $X\preceq^\B_\H X' \subset Y$, $X'\not\models P^Y$, 
(since $(X,Y)$ is a witness against $P \subseteq_{\t{\H,\B}} Q$),
and by the fact that $X\preceq^\B_\H Z$.

Only-if direction: W.l.o.g.\ assume $(X,Y)\in\sigma_{\t{\H,\B}}(P)\setminus
\sigma_{\t{\H,\B}}(Q)$; again, the other case is symmetric.
From $(X,Y)\in\sigma_{\t{\H,\B}}(P)$,
$(Y,Y)\in\sigma_{\t{\H,\B}}(P)$ follows by Definition~\ref{def:model}.
First consider, 
$(Y,Y)\notin\sigma_{\t{\H,\B}}(Q)$. 
Then either $Y\not\models Q$ or there 
exists an interpretation $Y'\subset Y$ with 
$Y'|_\H=Y|_\H$, such that $Y'\models Q^Y$.
In the former case $(Y,Y)$ is a witness against 
$P \subseteq_{\t{\H,\B}} Q$
and in the latter case $(Y',Y)$ is.
By
Lemma~\ref{lemma:witness}, $P \subseteq_{\t{\H,\B}} Q$ does not hold
and, consequently, $P \equiv_{\t{\H,\B}} Q$ does not hold as well.
Thus, let $X\subset Y$
and $(Y,Y)\in\sigma_{\t{\H,\B}}(Q)$.
We distinguish between two cases:

First suppose there exists an $X'$
with $X\preceq^\B_\H X'\subset Y$, such 
that $(X',Y)\in\sigma_{\t{\H,\B}}(Q)$.
Since $(X,Y)\notin\sigma_{\t{\H,\B}}(Q)$,
by definition of $\t{\H,\B}$-models, $X\prec^\B_\H X'$ has to hold,
and there exists a $Z\subset Y$ with $Z|_{\H\cup\B}=X'$, 
such that $Z\models Q^Y$. Note that $X\prec^\B_\H Z$ then also holds.
We show that 
$(Z,Y)$ is a witness against 
$P\subseteq_{\t{\H,\B}} Q$.
We already know that $Y$ is $\H$-total for $P$.
Moreover, 
we know $Z\models Q^Y$,
and
since $(X,Y)\in\sigma_{\t{\H,\B}}(P)$, 
we get by definition of $\t{\H,\B}$-models,
that,
for each $X''$ with
$X\prec^\B_\H X''\subset Y$,
$X''\not\models P^Y$.
Now since
$X\prec^\B_\H Z$,
Property (ii) in Definition~\ref{def:witness} holds for $(Z,Y)$.
This shows that
$(Z,Y)$ is 
a witness against $P\subseteq_{\t{\H,\B}} Q$.

So suppose, 
for each $X'$ with
$X\preceq^\B_\H X'\subset Y$, 
$(X',Y)\notin\sigma_{\t{\H,\B}}(Q)$ holds.
We have $(X,Y)\in\sigma_{\t{\H,\B}}(P)$, thus
there exists a $Z\subset Y$, with $Z|_{\H\cup \B}=X$, such that
$Z\models P^Y$.
We show that $(Z,Y)$ 
is a witness against the reverse 
problem,
$Q\subseteq_{\t{\H,\B}} P$.
From 
$(Y,Y)\in\sigma_{\t{\H,\B}}(Q)$, 
we get that 
$Y$ is an $\H$-total model of $Q$.
Moreover, we have $Z\models P^Y$. It 
remains
to show that, for each 
$X''$ with $Z\preceq^\B_\H X''\subset Y$,
$X''\not\models Q^Y$.
This holds by the assumption,
that
for each $X'$ with $X\preceq^\B_\H X'\subset Y$,
$(X',Y)\notin\sigma_{\t{\H,\B}}(Q)$, 
together with the fact that $Z|_{\H\cup \B}=X$.
Hence, 
both cases yield a witness, either 
against
$P\subseteq_{\t{\H,\B}} Q$ or
against
$Q\subseteq_{\t{\H,\B}} P$. By Lemma~\ref{lemma:witness} and 
Proposition~\ref{prop:eqcont}, 
$P\equiv_{\t{\H,\B}} Q$ does not hold.
\end{proof}

\begin{example}\label{ex:3}
In Example~\ref{ex:1}, we already mentioned that
$$
P= \{ a \vee b \la;\;a \la b \}
\quad\mbox{and}\quad
Q = \{ a \la \naf b;\; b\la\naf a;\; a \la b \}
$$
are
$\t{\{a,b\},\{b\}}$-equivalent.
Hence, fix $\H=\{a,b\}$, $\B=\{b\}$, and let us compute the
$\t{\H,\B}$-models of $P$, and resp., $Q$.
In Example~\ref{ex:2}
we already have 
seen that $Y_1=\{a,b\}$ and $Y_2=\{a\}$ 
are the models of both $P$ and $Q$. 
Since $\H=\{a,b\}$, 
both are $\H$-total models for $P$ and $Q$.
So,
$(Y_1,Y_1)$ and $(Y_2,Y_2)$ are the total
$\t{\H,\B}$-models of both programs.
It remains to check whether the non-total $\t{\H,\B}$-models of $P$ and $Q$
coincide.
First observe that $(Y_2,Y_1)$ is a $\t{\H,\B}$-model of both $P$ and $Q$, 
as well.
The interesting candidate is $(\emptyset,Y_1)$ since
$\emptyset$ is model of $Q^{Y_1}$ but not of $P^{Y_1}$. 
Hence, $(\emptyset,Y_1)$ cannot 
be $\t{\H,\B}$-model of $P$. But $(\emptyset,Y_1)$ is also not
$\t{\H,\B}$-model of $Q$, since there exists an interpretation
$X'$ satisfying 
$Q^{Y_1}$,
such that
$\emptyset \prec_\H^\B X' \subset Y$.
Take $X'=\{a\}$. Then, 
$\emptyset|_\H \subset X'|_\H$ and 
$X'_\B \subseteq \emptyset$ hold, 
which shows that $X'$ satisfies $\emptyset \prec_\H^\B X'$.
As another example, consider
$\H'=\{a\}$ and $\B'=\{a,b\}$. 
As we have seen in Example~\ref{ex:1}, $P\equiv_{\t{\H',B'}} Q$ holds, as well.
One can show that
$(Y_2,Y_2)$ is the only $\t{\H',\B'}$-model (over $\{a,b\}$) 
of $P$ as well as of $Q$.
This holds in particular, 
since $Y_1$ is neither an  $\H'$-total model 
of $P$ nor of $Q$ in this setting.

Let us also consider the parameterizations where $\t{\H,\B}$-equivalence between $P$ and $Q$
does not hold. 
For instance, this the case for $\H=\{b\}$,
$\B=\{a,b\}$ (see Example~\ref{ex:1} and~\ref{ex:2}).
We show that for
$Y = \{a,b\}$  and
$X=\emptyset$,
$(X,Y)$ is an $\t{\H,\B}$-model of $Q$ but not of $P$.
From Example~\ref{ex:2} we know that 
$Y$ is an $\H$-total model of $P$ as well as of $Q$.
Moreover,
$X\models Q^Y$ and $(X,Y)$ is $\preceq^\B_\H$-maximal for $Q$.
This is seen by the fact that the only $X'$, such that $X\prec^\B_\H X'\subset Y$ holds, is 
$X$ itself.  On the other hand, $X\not\models P^Y$, which is sufficient to 
see that $(X,Y)$ is not an $\t{\H,\B}$-model of $P$.
Also for $\H'=\{a,b\}$, $\B'=\{a\}$, we got $P\not\equivs_{\t{\H',\B'}} Q$.
Again it can 
be shown that
$(X,Y)$ with $X=\emptyset$ and $Y=\{a,b\}$ is an $\t{\H',\B'}$-model of $Q$ but not of $P$.
Now $\B'=\{a\}$, so in order to make $(X,Y)$ $\preceq^{\B'}_{\H'}$-maximal for $Q$, 
we also have to check that 
$X'\not\models Q^Y$,
for $X'=\{b\}$.
In fact, this is the case and thus $(X,Y)$ is an $\t{\H',\B'}$-model of $Q$. 
By the same observations as before, one shows that 
$(X,Y)$, in turn, 
is not $\t{\H',\B'}$-model of $P$.
Hence, the two cases where equivalence between $P$ and $Q$ does not hold, 
$P$ and $Q$ differ in their respective characterizations.
\hfill$\diamond$
\end{example}

\section{Special Cases}

In this section, we analyze the behavior of $\t{\H,B}$-models 
on special instantiations for $\H$ and $\B$.
We first consider the case where either $\H=\U$ or $\B=\U$.
We call the former scenario 
\emph{body-relativized}
and 
the latter
\emph{head-relativized}. 
Then, we sketch more general settings where the
only restriction is that either $\H\subseteq \B$ or $\B\subseteq \H$
holds.
The combination of the latter two is of particular interest since
it amounts to strong equivalence relative to $\H=\B$.

\subsection{Body-Relativized and Head-Relativized Equivalence}

First, we consider 
$\langle \U,\B \rangle$-equivalence problems,
in which $\U$ is fixed to be the universe, but
$\B$ can be arbitrarily chosen.
Note that $\langle \U,\B \rangle$-equivalence 
ranges from
strong 
(setting $\B=\U$) 
to uniform equivalence
(setting $\B=\emptyset$ and cf.\
Corollary~\ref{cor:unary}) 
and thus provides 
a common view on these two important problems, as well as 
on
problems ``in between'' them.
Second,
head-relativized equivalence problems,
$P\equivs_{\t{\H,\U}} Q$,
have as special cases 
once more 
strong equivalence (now by setting $\H=\U$)
but also the case where
$\H=\emptyset$ is of interest, since
it amounts to check whether $P$ and $Q$
possess the same answer sets under any addition of constraints.
It is quite obvious that
this holds iff
$P$ and $Q$ are ordinarily equivalent,
since
constraints can only ``rule out'' answer sets.
That observation is also reflected in Corollary~\ref{cor:unary}, 
since 
the only unary program in $\C_{\t{\emptyset,\U}}$ 
is the empty program.

The following result simplifies the definition of $\preceq_\H^\B$ 
within these settings.

\begin{proposition}
For interpretations $V,Z\subseteq\U$ and an alphabet $\A\subseteq \U$, 
it holds that
(i)~$V\preceq^\A_\U Z$ iff $V\subseteq Z$ and $V|_\A=Z|_\A$; and
(ii) $V\preceq^\U_\A Z$ iff $Z\subseteq V$ and $V|_\A=Z|_\A$.
\end{proposition}

Thus, maximizing w.r.t.\
$\preceq^\B_\H$ 
turns 
in the case of 
$\H=\U$
into
a form of $\subseteq$-maximization; and
in the case of 
$\B=\U$
into
a form of $\subseteq$-minimization.
Obviously, both neutralize themselves for $\B=\H=\U$, 
\iec in the strong equivalence setting,
where, by definition,
$V\preceq^\U_\U Z$ iff $V=Z$.

For body-relativized equivalence, our characterization now 
simplifies as follows.

\begin{corollary}\label{cor:seue}
A pair $(X,Y)$ of interpretations 
is an $\t{\U,\B}$-model of $P\in \CU$ iff
$X\subseteq Y$, 
$Y\models P$, $X\models P^Y$, and 
for all
$X'$ with $X\subset X'\subset Y$ and $X'|_\B=X|_\B$, 
$X'\not\models P^Y$.
\end{corollary}

Observe that for the notions in between strong  and uniform
equivalence the maximality test,
\iec checking
if each 
$X'$ with $X\subset X'\subset Y$ and $X'|_\B=X|_\B$ yields
$X'\not \models P^Y$,
gets more localized the more
atoms are contained in $\B$. In particular, for $\B=\U$ it 
disappears and we end up with a very simple condition
for $\t{\U,\U}$-models which exactly
matches the definition of SE-models by Turner~\shortcite{Turner03}:
a pair $(X,Y)$ of interpretations is an
SE-model  of a program $P$
iff 
$X\subseteq Y$, $Y\models P$, and $X\models P^Y$. 

For $\B=\emptyset$, 
on the other hand,
we observe that $X'|_\B=X|_\B$ 
always holds for $\B=\emptyset$.
Thus, 
a pair
$(X,Y)$ is a $\t{\U,\emptyset}$-model of a program $P$, if
$X\subseteq Y$,
$Y\models P$, $X\models P^Y$, and
for all
$X'$ with $X\subset X'\subset Y$, $X'\not\models P^Y$.
These conditions
are 
exactly 
the ones given for UE-models in~\cite{eite-fink-03}.
Hence, Corollary~\ref{cor:seue} provides a common view on 
the characterizations of
uniform and strong equivalence.

For head-relativized equivalence notions, 
simplifications are as follows.

\begin{corollary}\label{cor:he}
A pair $(X,Y)$ of interpretations 
is an 
$\t{\H,\U}$-model of $P\in\CU$ iff
$X\subseteq Y$,
$Y$ is an $\H$-total model for $P$, 
$X\models P^Y$, 
and,
for each 
$X'\subset X$ 
with $X'|_\H = X|_\H$,
$X'\not\models P^Y$.
\end{corollary}

In the case of $\H=\U$, 
$\t{\H,\U}$-models 
again
reduce to 
SE-models.
The other special case 
is $\H=\emptyset$.
Recall that $\t{\emptyset,\U}$-equivalence
amounts to ordinary equivalence.
$\t{\emptyset,\U}$-models thus characterize answer sets as follows:
First, 
observe that 
all
$\t{\emptyset,\U}$-models
have to be total.
Moreover, 
$(Y,Y)$ is an $\emptyset$-total model for $P$,
iff
no $X\subset Y$ satisfies $X\models P^Y$, 
\iec iff $Y$ is an answer set of $P$.
So there is
a one-to-one correspondence between
the
$\t{\emptyset,\U}$-models
and the
answer sets of a program.

\subsection{%
$\B\subseteq\H$ - and 
$\H\subseteq \B$ - Equivalence}

We just highlight a few results here, 
in order to establish a connection between $\t{\H,\B}$-models
and relativized SE- and UE-models, as defined by Woltran~\shortcite{Woltran04}.

\begin{proposition}\label{prop:subset}
For interpretations $V,Z\subseteq \U$ and alphabets $\H,\B\subseteq\U$ with 
$\B\subseteq \H$ (resp., 
$\H\subseteq \B$), 
$V\preceq_\H^\B Z$ iff 
$V|_\H\subseteq Z|_\H$ and $V|_\B=Z|_\B$ 
(resp., iff 
$Z|_\B\subseteq V|_\B$ and $V|_\H=Z|_\H$).  
Moreover, if $\A=\H=\B$,
$V\preceq_\H^\B Z$ iff 
$V|_\A = Z|_\A$.
\end{proposition}

Observe that $\preceq^\A_\A$-maximality 
(in the sense of Definition~\ref{def:max})
of a pair $(X,Y)$ for $P$  
reduces to test $X\models P^Y$.
Thus, to make $(X|_\A,Y)$ an $\t{\A,\A}$-model of $P$, we just 
additionally need 
$\A$-totality of $Y$.
In other words, we
obtain the following criteria.

\begin{corollary}\label{cor:bsubh}
Given $\A\subseteq\U$, 
a pair $(X,Y)$ of interpretations 
is an $\t{\A,\A}$-model of a program $P\in\CU$,
iff
(1) $X=Y$ or $X\subset Y|_\A$, 
(2) $Y\models P$ and for each $Y'\subset Y$, $Y'\models P^Y$ implies 
$Y'|_\A \subset Y|_\A$; and 
(3) if $X\subset Y$ then there exists an $X'\subseteq Y$ with $X'|_\A=X$, 
such that $X'\models P^Y$. 
\end{corollary}

This exactly matches the definition of $\A$-SE-models~\cite{Woltran04}. 
Finally, if we switch from 
$\t{\A,\A}$-equivalence to $\t{\A,\emptyset}$-equivalence 
(\iec from relativized strong to relativized uniform equivalence)
we obtain the following result for $\t{\A,\emptyset}$-models
which can be shown to coincide with the 
explicit definition of $\A$-UE-models~\cite{Woltran04}.

\begin{corollary}\label{cor:h}
Given $\A\subseteq\U$, 
a pair $(X,Y)$ of interpretations 
is an $\t{\A,\emptyset}$-model of $P\in\CU$,
iff
(1) and (2) from Corollary~\ref{cor:bsubh} hold, and 
if $X\subset Y$ then there exists $X'\subseteq Y$ 
such that
$X'|_\A=X$, 
$X'\models P^Y$,
and for each $X''\subset Y$ 
with $X \subset X''|_\A$,
$X''\not\models P^Y$.
\end{corollary}

Thus, the concept of $\t{\H,\B}$-models captures known 
characterizations from the literature, in particular, 
$\A$-SE-models and $\A$-UE-models~\cite{Woltran04}, 
which themselves include the prominent characterizations of
SE-models~\cite{Turner03} and 
UE-models~\cite{eite-fink-03} as a special case.

\section{Discussion}

In this section we 
first consider the case 
of positive programs, and show how the characterization for $\t{\H,\B}$-equivalence
simplifies for such programs.
Moreover, 
we address the computational complexity 
of checking $\t{\H,\B}$-equivalence. 
Finally, we informally discuss a method
for implementing a decision procedure for $\t{\H,\B}$-equivalence.


When comparing positive programs with respect to 
$\t{\H,\B}$-equivalence, it turns out that the actual 
parameterization for $\B$ is immaterial.

\begin{theorem}\label{thm:pos}
For any positive programs $P,Q\in\CU$ and alphabets $\H,\B\subseteq \U$, 
we have $P\equivs_{\t{\H,\B}} Q$ iff
$P$ and $Q$ possess the same $\H$-total models.
\end{theorem}
\begin{proof}
The only-if
direction is obvious, since if w.l.o.g.\ $Y$ is an $\H$-total model of $P$
but not of $Q$, we obtain immediately, 
$(Y,Y)\in\sigma_{\t{\H,\B}}(P)\setminus \sigma_{\t{\H,\B}}(Q)$ and thus
by Theorem~\ref{thm:central}, $P\not\equivs_{\t{\H,\B}} Q$.

For the only if-direction we get from 
$P\not\equivs_{\t{\H,\B}} Q$
that the $\t{\H,\B}$-models  of $P$ and $Q$ differ, 
again using Theorem~\ref{thm:central}. W.l.o.g.\ assume a pair
$(X,Y)$ such that $(X,Y)\in\sigma_{\t{\H,\B}}(P)\setminus \sigma_{\t{\H,\B}}(Q)$. 
If $X=Y$ the $\H$-total models of $P$ and $Q$ differ by definition.
So suppose $P$ and $Q$ have the same total $\t{\H,\B}$-models.
\iec $X\subset Y$ holds.
Hence, we have $(Y,Y)\in\sigma_{\t{\H,\B}}(P)$ 
(otherwise $(X,Y)\in\sigma_{\t{\H,\B}}(P)$ would not hold)
and also $(Y,Y)\in\sigma_{\t{\H,\B}}(Q)$.
By the latter, we have 
two reasons remaining for
$(X,Y)\notin\sigma_{\t{\H,\B}}(Q)$:
(i) no $X'$ with $X'\subset Y$ and $X'|_{\H\cup \B} = X$ satisfies $Q^Y=Q$;
(ii) there exists an $X'$ with $X\prec^\B_\H X'\subset Y$, such 
that $X'\models Q^Y=Q$. 
Also recall that 
since $(X,Y)\in\sigma_{\t{\H,\B}}(P)$,
there exists an $X'\subset Y$ with $X'|_{\H\cup \B} = X$, such 
that $(X',Y)$ is $\preceq^\B_\H$-maximal for $P$.
For Case (ii)  we thus
get that there exists an 
$\t{\H,\B}$-model $(X'',Y)$ of $Q$, 
such that $X\prec^\B_\H X''$.
But $(X'',Y)$ cannot be an $\t{\H,\B}$-model of $P$, since
$(X',Y)$ is already $\preceq^\B_\H$-maximal for $P$.
Hence, we have the same situation as Case~(i) with $P$ and $Q$ exchanged.

So it remains to  discuss Case (i). Let
$(X',Y)$ be $\preceq^\B_\H$-maximal for $P$ as above, and
let us w.l.o.g.\ select the subset-minimal interpretation $X'$ with 
$X'|_{\H\cup \B} = X$
which is $\preceq^\B_\H$-maximal for $P$.
Clearly, $X'\models P^Y=P$
and we show that $X'$ is an $\H$-total model of $P$.
Towards a contradiction suppose this is not the case, \iec there
exists an $X''\subset X'$ with $X''|_\H=X'|_\H$, 
such that $X''\models P^{X'}=P$. 
Observe that $X''|_\B = X'|_\B$ 
cannot be the case, since we selected $X'$ as the minimal 
interpretation which satisfies $P$ such that $X'|_{\H\cup \B} = X$ holds.
Hence, $X''|_\B\subset X'_\B$ has to hold, but then, by definition
$X'\prec^\B_\H X''$ would hold and thus 
$(X',Y)$ would not be  $\preceq^\B_\H$-maximal for $P$, as assumed.
On the other hand, $X'$ cannot be an $\H$-total model of $Q$, since 
$X'$ is not even a model of $Q$ in view of the assumption for 
Case~(i).
\end{proof}

We proceed by providing complexity results
of the decision problem for $\t{\H,\B}$-equivalence.

\begin{theorem}
Given program $P,Q\in\CU$ and alphabets $\H,\B\subseteq\U$, 
deciding $P\equivs_{\t{\H,\B}} Q$ is $\PiP{2}$-complete; $\PiP{2}$-hardness 
holds even for positive programs.
Deciding $P\equivs_{\t{\H,\B}} Q$ is $\CONP$-complete if $P$ and $Q$ are normal programs.
\end{theorem}
\begin{proof}
Former results on 
relativized
strong equivalence~\cite{Eiter07} 
show that
the problem is $\PiP{2}$-hard even for positive disjunctive logic programs, 
and $\CONP$-hard for normal logic programs.
Since relativized strong equivalence is a special case of 
$\t{\H,\B}$-equivalence, these lower bounds hold for 
$\t{\H,\B}$-equivalence, as well.
The corresponding membership  results
hold in view of 
Corollary~\ref{cor:unary}. 
In particular, 
one can guess an interpretation $Y$
and a unary program $R\in\C_{\t{\H,\B}}$, 
and then check whether $Y$ is contained in either 
$\AS(P\cup R)$ or 
$\AS(Q\cup R)$, but not in both. Answer-set checking is in $\CONP$ in general, 
and in $\Pol$ for normal logic programs.
Since one can safely restrict $Y$ and $R$ 
to contain only atoms which also 
occur in $P$ or $Q$, 
this algorithm
for disproving
$\t{\H,\B}$-equivalence
runs in nondeterministic polynomial time 
for normal programs, resp.\ 
in nondeterministic polynomial time
with access to an $\NP$-oracle for the general case of disjunctive programs.
Thus, that problem is in $\NP$ (resp., in $\SigmaP{2}$), and consequently
$\t{\H,\B}$-equivalence is in 
$\CONP$ for normal programs and in 
$\PiP{2}$, in general.
\end{proof}

The 
complexity results we obtained show
that $\t{\H,\B}$-equivalence can be efficiently reduced, for instance, to 
ordinary equivalence, such that the class of programs is retained.
We briefly discuss an 
approach which makes use of Corollary~\ref{cor:unary} in a similar manner
as in above membership proof
and compiles $\t{\H,\B}$-equivalence into ordinary equivalence
for which a dedicated system exists~\cite{Janhunen04a};
a similar method was also discussed 
in~\cite{Woltran04,OJ06:ecai}.
The idea hereby is to
incorporate the
guess of the unary context programs over the specified
alphabets in both programs accordingly. 
To this end, let,
for an  $\t{\H,\B}$-equivalence problem between programs $P$ and $Q$, 
$f$ 
as well as 
$c_{a,b}$ and $\bar{c}_{a,b}$ for each $a\in \H$, $b\in \B\cup \{f\}$,
be new distinct atoms, not occurring in $P\cup Q$. 
Moreover, let 
$$
R_{\t{\H,B}} = 
\Big\{ c_{a,b}\vee \bar{c}_{a,b}\la;\;
a \la b, c_{a,b}\mid a\in \H,b\in \B\cup \{f\}  \Big\}
\cup 
\{ f\la \}
$$
which is used to guess a context program.
In fact, the role of atoms $c_{a,f}$ is to 
guess a set of facts $F\subseteq \H$, 
while atoms $c_{a,b}$ with $b\neq f$ guess a 
subset of 
unary rules $a\la b$ 
with $a\in\H$ and $b\in\B$.

Then, $P\equiv_{\t{\H,B}} Q$ holds iff
$P \cup R_{\t{\H,B}}$ 
and 
$Q \cup R_{\t{\H,B}}$ 
are ordinarily equivalent; showing this correspondence is rather straightforward, in particular
by application of the splitting theorem~\cite{Lifschitz94}.
Note that $P \cup R_{\t{\H,B}}$ and $Q \cup R_{\t{\H,B}}$ are positive whenever 
$P$ and $Q$ are positive.
Moreover, we can replace
in $R_{\t{\H,B}}$ each disjunctive facts
$c_{a,b}\vee \bar{c}_{a,b}\la$ by 
two corresponding normal rules 
$c_{a,b}\la \naf \bar{c}_{a,b}$ and
$\bar{c}_{a,b}\la \naf c_{a,b}$.
Hence, if we want to decide $\t{\H,\B}$-equivalence between two normal programs, our
method results in an ordinary equivalence problem between normal programs, as well.

\section{Conclusion}

The aim of this work is to provide a general and uniform characterization
for different equivalence problems, which have been 
handled
by inherently different concepts, so far.
To this end, 
we have introduced an equivalence notion
parameterized by two alphabets to restrict the 
atoms allowed to occur in the heads, and respectively, bodies
of the 
context programs. 
We showed that
our approach captures the most important equivalence
notions 
studied, including strong and uniform equivalence
as well as 
relativized notions thereof.

\begin{figure}[t]
\begin{center}
\begin{picture}(200,100)(0,0)
\thicklines
\put(25,10){\vector(1,0){145}}
\put(170,10){\vector(0,1){73}}
\put(25,10){\vector(2,1){145}}
\put(25,10){\vector(0,1){73}}
\put(25,83){\vector(1,0){145}}
\put(0,0){$\t{\emptyset,\emptyset}$}
\put(70,0){\makebox(70,0){ordinary equivalence}}
\put(70,90){\makebox(70,0){BRE}}
\put(14,40){\begin{rotate}{90} RUE \end{rotate}}
\put(180,40){\begin{rotate}{90} HRE \end{rotate}}
\put(60,35){\begin{rotate}{25} RSE \end{rotate}}
\put(40,70){$\B\subseteq\H$}
\put(135,25){$\H\subseteq\B$}
\put(0,90){$\t{\U,\emptyset}$=UE}
\put(170,0){$\t{\emptyset,\U}$}
\put(170,90){SE=$\t{\U,\U}$}
\end{picture}
\end{center}
\caption{The landscape of $\t{\H,\B}$-equivalence with either $\H\subseteq \B$ or $\B\subseteq \H$.}
\label{fig:1}
\end{figure}

Figure~\ref{fig:1} 
gives an overview of $\t{\H,\B}$-equivalence and its special cases, 
\iec 
relativized uniform equivalence (RUE),
relativized strong equivalence (RSE),
body-relativ\-ized equivalence (BRE),
and head-relativized equivalence (HRE).
On the bottom line we have ordinary equivalence, while the top-left
corner
amounts to uniform equivalence (UE) and the 
top-right corner to strong
equivalence (SE).

Future work includes the study of further properties 
of $\t{\H,\B}$-equivalence, 
as well as potential applications, which include relations
to 
open logic programs \cite{Bonatti:2001:ROL}
and 
new concepts for 
program simplification~\cite{Eiter03a}.
An extension of 
$\t{\H,\B}$-equivalence from disjunctive logic programs to theories 
is a further aspect to be considered. In particular, this requires a 
reformulation of the concept of $\t{\H,\B}$-models in terms of the logic 
of here-and-there (which was used in \cite{Lifschitz01} to characterize strong
equivalence between theories under equilibrium logic, a generalization of the answer-set semantics).
We expect that such a generalization of our results can be accomplished in similar manner
as this was done for 
relativized strong and uniform equivalence~\cite{Pearce07a}.
Also an extension of the framework 
in the sense of~\cite{Eiter05}, where a 
further alphabet 
for answer-set projection 
is used to specify the atoms which have to coincide
in comparing the answer sets is of interest.
While Eiter et al.~\shortcite{Eiter05} provide
a characterization for relativized \emph{strong} equivalence
with projection, recent work \cite{Oetsch07b} addresses the 
problem of relativized \emph{uniform} equivalence with projection.
Our results may be a basis to provide a common
view on these two 
concepts of program correspondence, as well.

\subsubsection*{Acknowledgments.}
This work was 
supported 
by the Austrian Science Fund (FWF)
under grant P18019. 
I would like thank the anonymous reviewers for useful comments.

\end{document}